\documentclass[10pt]{article}
\usepackage{tommoro-report}
\usepackage{wrapfig}
\usepackage{float}
\usepackage{placeins}
\usepackage{kotex}
\usepackage{amssymb}
\usepackage{amsmath}

\settrtitle{Habilis-$\boldsymbol{\beta}$: A Fast-Motion and Long-Lasting On-Device Vision-Language-Action Model}
\begin{document}

\begin{figure}[h]
    \centering
    \includegraphics[width=\linewidth]{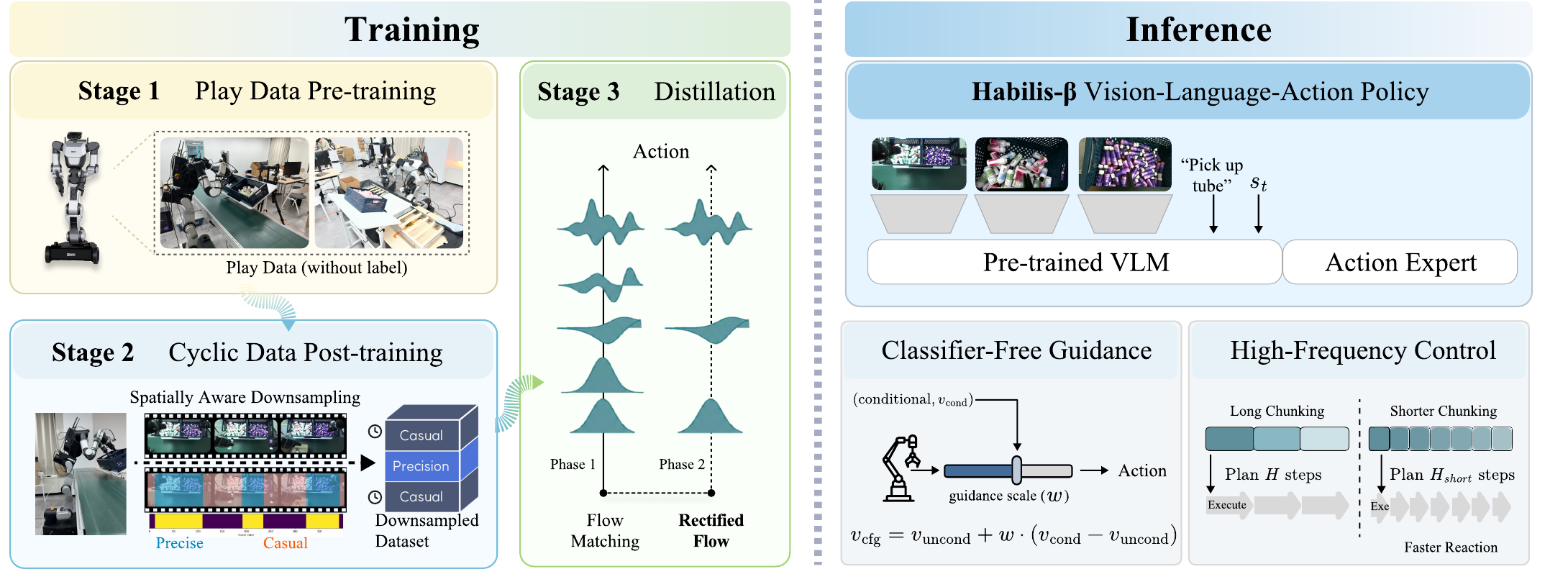}
    \caption{\textbf{Habilis-$\boldsymbol{\beta}$ Overview.} \textit{(Left)} \textbf{Training Pipeline:} The model is trained in three stages to achieve Fast-Motion and Long-Lasting capabilities. Stage 1 learns a robust, task-agnostic interaction prior via play data pre-training. Stage 2 post-trains on cyclic task demonstrations, utilizing Spatially Aware Downsampling (ESPADA) to compress casual free-space motions. Stage 3 distills the multi-step flow matching action expert into an efficient rectified flow model. \textit{(Right)} \textbf{Inference Pipeline:} To enable On-Device operation, a pre-trained VLM prefix fuses multimodal observations to condition the distilled action expert. The reduced inference cost is reinvested into High-Frequency Control, using shorter action chunks for rapid closed-loop reactivity. Finally, Classifier-Free Guidance (CFG) acts as a deployment-time knob to dynamically balance instruction adherence and learned interaction priors.}
    \label{fig:main_overview}
\end{figure}

\begin{trabstract}
We introduce Habilis-$\beta$, a fast-motion and long-lasting on-device vision-language-action (VLA) model designed for real-world deployment. Current VLA evaluation remains largely confined to single-trial success rates under curated resets, which fails to capture the fast-motion and long-lasting capabilities essential for practical operation. To address this, we introduce the Productivity–Reliability Plane (PRP), which evaluates performance through Tasks per Hour (TPH) and Mean Time Between Intervention (MTBI) under a continuous-run protocol that demands both high-speed execution and sustained robustness. 
Habilis-$\beta$ achieves high-performance by integrating language-free pre-training on large-scale play data for robust interaction priors with post-training on cyclic demonstrations that capture state drift across consecutive task iterations. The system further employs ESPADA for phase-adaptive motion shaping to accelerate free-space transit, utilizes rectified-flow distillation to enable high-frequency control on edge devices, and incorporates classifier-free guidance (CFG) as a deployment-time knob to dynamically balance instruction adherence and learned interaction priors. 
In 1-hour continuous-run evaluations, Habilis-$\beta$ achieves strong performance under the PRP metrics, compared to $\pi_{0.5}$ in both simulation and real-world environments. In simulation, Habilis-$\beta$ achieves $572.6$ TPH and $39.2$s MTBI (vs. $120.5$ TPH and $30.5$s for $\pi_{0.5}$), while in a real-world humanoid logistics workflow it achieves $124$ TPH and $137.4$s MTBI (vs. $19$ TPH and $46.1$s for $\pi_{0.5}$). Finally, Habilis-$\beta$ achieves the highest reported performance on the standard RoboTwin 2.0 leaderboard across representative tasks, validating its effectiveness in complex manipulation scenarios.
\end{trabstract}

\section{Introduction}
Vision-language-action (VLA) models \cite{clip-rt, pi05, openvla-oft, grootn15, octo2024, pmlr-v229-zitkovich23a, Brohan-RSS-23} have advanced rapidly. 
Motivated by generalization in language and vision~\cite{openai2024gpt4technicalreport}, recent work asks whether a similar generalization can be achieved in physical robot control.

However, several obstacles still prevent general VLA systems from working reliably in the real world. 
First, many VLA systems move slowly \cite{kim2025espada, guo2025demospeedup, arachchige2025sail}. 
They produce conservative, temporally dense actions that underuse fast free-space motion and show limited closed-loop reactivity, reducing throughput. 
Second, evaluation protocols still emphasize single-attempt trials under curated reset conditions \cite{kressgazit2024policy_eval_best_practices}. 
This hides what happens in extended operation, where tasks repeat, initial states drift, and small execution errors accumulate until the system fails and requires intervention.
Third, end-to-end latency remains a practical bottleneck \cite{williams2025litevla, budzianowski2025edgevla, clip-rt}. 
Large VLA stacks can require heavy computing and may rely on networked components, adding delay and reducing predictability when a fast reaction is necessary. 
These limitations matter for generality, and they are especially costly in industrial settings where productivity, reliability, and operating cost are hard constraints.

Accordingly, \textbf{Habilis-$\boldsymbol{\beta}$} focuses on three deployment requirements: Fast-Motion, Long-Lasting, and On-Device.
\textit{Fast-Motion} means maintaining high speed when precision is not critical, while switching to careful control near contact and during fine manipulation.
\textit{Long-Lasting} means sustaining performance under repeated execution as initial states drift and small errors accumulate, with minimal human intervention or resets.
\textit{On-Device} means running the perception-to-action loop locally on an edge device while preserving Fast-Motion and Long-Lasting performance, enabling predictable latency and reduced dependence on network or service variability under tight compute and power budgets.

Meeting these requirements requires evaluation protocols that reflect repeated execution and real-time constraints.
We revisit a common assumption in VLA training and evaluation: episodes are treated as independent, single-trial trajectories \cite{kressgazit2024policy_eval_best_practices}. 
Under this ``non-cyclic'' regime, the success rate measures how often a model completes a task once from a carefully prepared initial state. 
However, we further argue that the single-trial success rate is not a sufficient deployment metric. 
It collapses the speed–accuracy trade-off and qualitatively different failure modes into one number, and it is insensitive to long-run degradation and intervention frequency.
In deployment, failures come in many forms (e.g., timeouts, aborts) and they do not have a single clean definition.
What matters operationally is whether the run can continue without external help \cite{buerkle2023iraas}.

We therefore propose a Productivity-Reliability Plane (PRP) where we employ Tasks per Hour (TPH) and Mean Time Between Intervention (MTBI) as our metrics. 
TPH measures sustained throughput under realistic operating constraints and serves as a primary measure of Fast-Motion capability. 
MTBI measures the average operating time between events that require external intervention and serves as a primary measure of Long-Lasting capability. 
To make long-duration evaluation well-defined, we treat any such event that requires model-external assistance (e.g., reset or recovery) as an \textit{intervention}.
Considering both metrics jointly exposes the productivity-reliability trade-off that dominates real deployments and distinguishes systems that otherwise look similar under the single-trial success rate.

These deployment metrics also reshape what we want from data.
In our early experiments, we find that models trained under standard episode segmentation can achieve a high single-trial success rate, yet deteriorate rapidly under repetition \cite{laskey2017dart, gupta2021resetfree}.
A plausible driver is distribution shift: repeated execution induces drifting initial states and recovery situations that non-cyclic data rarely contain \cite{yoon2026how, paudel2022learningrobotdecisionmaking}.
We therefore leverage \textit{play} data and \textit{cyclic} task data as complementary mechanisms to broaden coverage and better match the state distribution encountered in continuous operation.

We leverage play data \cite{lynch2020learninglatentplansfromplay, LynchSermanet2021LCIL} as a scalable way to learn a broad interaction prior grounded in physical behavior.
Play data refers to unstructured, unsegmented demonstration data collected from a human teleoperating a robot without any explicit task labels or goal specifications.
Because it is relatively cheap to collect, we use it for language-unconditioned pre-training to learn a task-agnostic motion prior.

We then post-train on task demonstrations using a ``cyclic'' dataset formulation that better matches the state distribution encountered in continuous operation.
Concretely, instead of segmenting data as success-then-stop episodes, cyclic data records continuous trajectories where the robot repeatedly executes the same instruction across many cycles, linking the end of one success to the start of the next.
Training on cyclic data encourages the model to learn recovery behavior and to retain performance over long runs, which are central to deployment.

The same mismatch appears on the productivity side. 
When evaluation focuses on one-shot success, datasets and policies have little incentive to learn when fast-motion is appropriate, learned behaviors often remain temporally dense and overly conservative. 
We address this by shaping demonstrations so that free-space motion can be executed quickly, while contact-rich phases remain precise and reactive. 
Specifically, we apply ESPADA \cite{kim2025espada}, our recent method that compresses redundant action sequences and prioritizes high-impact interactions. 
This reduces unnecessary temporal density in the learned behavior and mitigates slow motion, improving throughput without sacrificing task success.

Additionally, to achieve end-to-end responsiveness, we combine rectified flow distillation \cite{liu2022rectified_flow} with high-frequency control. 
Distillation reduces the inference cost and frees the latency budget under on-device or local-machine constraints. 
Rather than spending this budget solely on a single slow pass, we allocate it to more frequent re-inference. 
This improves closed-loop reactivity and enables rapid adaptation to unexpected events while keeping latency predictable.

Finally, we optionally apply classifier-free guidance (CFG) \cite{ho2022cfg}, interpolating between instruction-conditioned and instruction-free predictions.
In practice, increasing guidance strengthens instruction adherence, shifting the policy toward more proactive behavior and often increasing execution speed and behavioral aggressiveness.
CFG therefore serves as a convenient deployment-time knob for navigating the speed–reliability trade-off.

For the evaluation, we build a continuous-run environment in both simulation and the real world. 
Instead of single-trial episodes, we run policies in long-horizon autonomous cycles where tasks repeat back-to-back, initial states drift, and occasional interventions become necessary. 
This makes the productivity–reliability trade-off explicit and yields stable measurements of TPH, MTBI, and success rate under sustained operation.

In this continuous-run regime, Habilis-$\beta$ improves all three deployment metrics.
Overall, Habilis-$\beta$ substantially outperforms the strongest baseline $\pi_{0.5}$ in both simulation and the real world under long-horizon operation.
In simulation, Habilis-$\beta$ achieves $572.6$ TPH, $39.2$s MTBI, and a $78.5\%$ success rate, compared to $\pi_{0.5}$ at $120.5$ TPH, $30.5$s MTBI, and a $47.8\%$ success rate.
In the real world, under a 1-hour continuous-run protocol, Habilis-$\beta$ achieves $124$ TPH, $137.4$s MTBI, and an $82.7\%$ success rate, compared to $\pi_{0.5}$ at $19$ TPH, $46.1$s MTBI, and a $19.6\%$ success rate.

We further benchmark on the RoboTwin 2.0 \cite{chen2025robotwin} suite using its standard single-trial evaluation protocol and report results on three representative tasks: \texttt{Dump Bin Bigbin}, \texttt{Place Dual Shoes}, and \texttt{Stack Bowls Three}. 
Habilis-$\beta$ records the best reported performance on these RoboTwin 2.0 tasks at the time of release.

\section{Problem Setup and Metrics}
We evaluate deployment readiness along two operational axes: \textit{productivity} (how much work it completes per unit time under the same deployment protocol) and \textit{reliability} (how long the system can run without human intervention).

\subsection{Deployment Protocol: Continuous-Run Evaluation}
We use a continuous-run protocol that mirrors production operation. 
The robot repeatedly executes a fixed task cycle for a fixed wall-clock duration, while the environment is re-initialized only when necessary. 
For each model/setting, we run the system for a fixed duration $T$ (typically $T=3600$ seconds). 
During the run, the robot continuously attempts task cycles. We record interventions whenever the execution cannot proceed without external human assistance, after which the run resumes (e.g., via reset/restart). 
In our experiments, interventions are defined by two mechanisms. 
First, a \textit{timeout rule}: if the robot fails to complete a task cycle within a task-specific time limit, we record an intervention and reset. 
Second, an \textit{abort} (real-world only): any operator-initiated termination (e.g., an E-stop) that requires external recovery is recorded as an intervention.
\footnote{We use task-specific timeouts to avoid unfairly penalizing tasks with inherently different nominal cycle times. Exact timeouts are reported per task in Section~\ref{sec:experiments}.}
This protocol emphasizes performance retention under repetition and makes compounding drift, stuck accumulation, and recovery failures measurable.

\newpage

\subsection{Productivity Metric: Tasks per Hour (TPH)}
Productivity is measured as TPH, i.e., the number of completed tasks per hour under the same continuous-run protocol.
Let $N_{\mathrm{succ}}$ be the number of successful task completions during the run. 
Our primary reporting uses wall-clock normalization:
\begin{equation}
\mathrm{TPH} \;=\; \frac{N_{\mathrm{succ}}}{T/3600}.
\end{equation}

\subsection{Reliability Metric: Mean Time Between Intervention (MTBI)}
Reliability is quantified by Mean Time Between Intervention (MTBI), defined as the average continuous run time between required external interventions.
Let $K$ be the number of interventions during a continuous run of duration $T$ seconds.
We compute
\begin{equation}
\mathrm{MTBI} \;=\; \frac{T}{K}.
\end{equation}
Under the standard one-hour protocol ($T=3600$), this simplifies to $\mathrm{MTBI} = 3600/K$.

\subsection{Productivity-Reliability Plane}
To evaluate productivity and reliability jointly, we visualize each model/setting as a point on a 2D plane:
\begin{itemize}
  \item x-axis: $\mathrm{TPH}$ (productivity),
  \item y-axis: $\mathrm{MTBI}$ (reliability).
\end{itemize}
We refer to this visualization as the Productivity-Reliability Plane (PRP). 
A method improves deployment readiness if it achieves higher MTBI and higher TPH under the same continuous-run protocol.

\section{Habilis-$\beta$: Fast-Motion, Long-Lasting, On-Device VLA}

\subsection{System Overview}
\begin{figure}[h]
    \centering
    \includegraphics[width=0.9\linewidth]{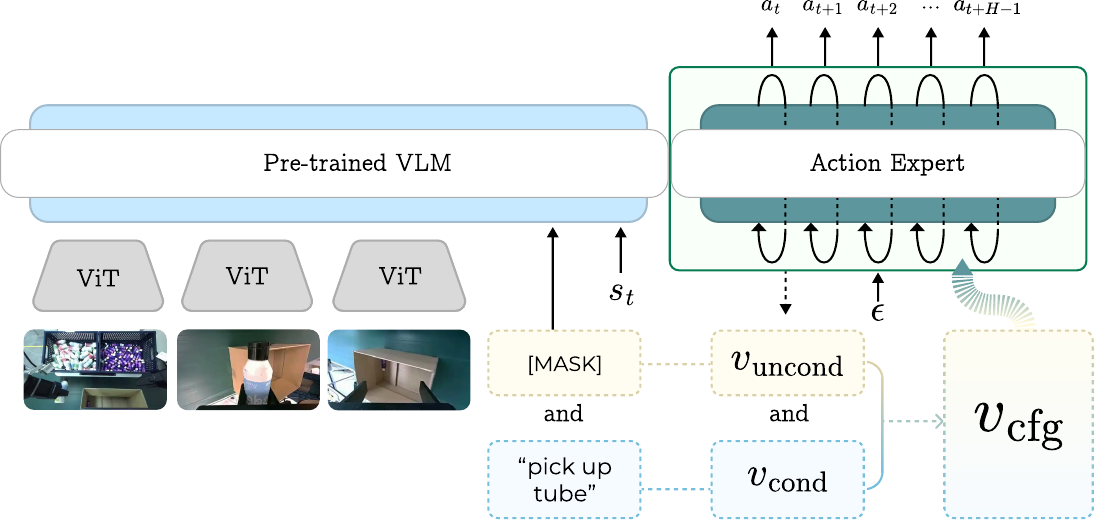}
    \caption{\textbf{Model Architecture.} Habilis-$\beta$'s prefix-suffix architecture uses a pre-trained VLM prefix to process multimodal inputs, conditioning a suffix action expert that generates continuous action chunks.}
    \label{fig:model_architecture}
\end{figure}

Figure~\ref{fig:model_architecture} illustrates the model architecture of Habilis-$\beta$.
The model takes multi-view observations and a high-level instruction as input and outputs H-step continuous action chunk $(a_{t}, \cdots, a_{t + H - 1})$ \cite{zhao2023aloha, liu2025bid}. 
Each camera view is embedded by a ViT \cite{dosovitskiy2021vit} encoder, and the resulting visual tokens are fused by a pre-trained VLM backbone.
An action expert then consumes the proprioception $s_{t}$ and generates the action chunk. 
At inference time, we optionally apply CFG \cite{ho2022cfg} by mixing an instruction-conditioned prediction and an instruction-masked prediction.

\subsection{Model Architecture and Training}
\label{sec:training}
\subsubsection{Flow Matching Action Expert}
The action expert in Habilis-$\beta$ uses flow matching \cite{lipman2023flow_matching} to model continuous action chunks. 
At an environment time $t$, an observation is given by $o_{t}=(I_{t},s_{t},\ell)$ where $I_{t}$ is a set of images, $s_{t}$ is a state, and $\ell$ is an instruction. 
We define a linear interpolation (also called noisy action) $x_{\tau} $ between a demonstrated action chunk $a \in \mathbb{R}^{H \times d}$ and Gaussian noise $\epsilon \in \mathbb{R}^{H \times d}$ with action horizon $H$ and action dimension $d$: 
\begin{equation}
    \epsilon \sim \mathcal{N}(0, I), \qquad
    \tau \sim \mathrm{Beta}(1.5, 1), \qquad
    \tau \leftarrow 0.999\,\tau + 0.001,
\end{equation}
\begin{equation}
    x_{\tau} = \tau\,\epsilon + (1 - \tau)\,a, 
    \qquad
    u_{\tau} = \frac{dx_{\tau}}{d\tau} = \epsilon - a.
\end{equation}
Here, $\tau$ is denoising time sampled from a Beta distribution $\mathrm{Beta}(1.5, 1)$ affinely mapped to $[0.001, 1.0]$.
The model predicts a velocity field $v_{\theta} (o, x_{\tau}, \tau)$ over the entire horizon $H$ where $\theta$ denotes the learnable parameters.
It is trained to match the target velocity field $u_{\tau}$. 
We follow a prefix-suffix design \cite{pi2024pizero} where a pre-trained VLM processes multimodal inputs and the action expert is a suffix-conditioned decoder.
To enable controllable guidance at inference, we apply language-only unconditioning during training. 
The model is trained using flow matching regression with a CFG-style masking mechanism:
\begin{equation}
    \mathcal{L}_{\mathrm{FM}}(\theta)
    = \mathbb{E}_{\epsilon, \tau, (o,a) \sim \mathcal{D}}
    \left[
    \frac{1}{Hd}\left \Vert v_\theta(\tilde{o}, x_{\tau}, \tau) - u_{\tau} \right \Vert_2^2
    \right],
\end{equation}
where $\mathcal{D}$ is the dataset of trajectory-language pairs. 
Here, $\tilde{o}$ refers to the observation with stochastically masked instructions for learning unconditional dynamics.
This learning objective encourages the model to separate instruction-induced bias (task selection) from the low-level action prior that supports stable interaction.

\subsubsection{Rectified Flow Distillation}
To minimize inference latency without degrading generation quality, we adopt the \textbf{Rectified Flow}~\cite{liu2022rectified_flow, liu2024instaflow, zhu2025slimflow, esser2024scalingrectifiedflowtransformers}. 
Specifically, we leverage its distillation \cite{hinton2015distilling} capability to compress a multi($\geq10$)-step teacher model to a two-step student model. 
In our implementation, the teacher model $v_{\theta} (o_{t}, x_{\tau}, \tau)$ with a learnable parameter $\theta$ is a transformer that generates an action chunk through iterative ODE integration from noise to data.
The student model $v_\phi(o_{t}, x_{\tau}, \tau)$ with a learnable parameter $\phi$ is trained to bypass this iterative process by learning to predict the teacher's terminal target-induced velocity.
The distillation objective is defined as:
\begin{equation}
    \mathcal{L}_{\mathrm{distill}}(\phi) = \mathbb{E}_{o_{t} \sim \mathcal{D}, \tau \sim \mathcal{U}(0,1), x_{\tau} \sim p_{\theta}(\cdot \mid o_{t}, \tau)} \left[ \Vert v_\phi(o_{t}, x_{\tau}, \tau) - \frac{x_{1} - x_{\tau}}{1 - \tau} \Vert^{2} \right],
\end{equation}
$o_{t}$ is the multimodal observation context at environment time $t$, processed by the VLM prefix.
$x_{\tau}$ is an intermediate state at denoising step $\tau$ along the teacher's ODE trajectory.
$x_{1}$ is the final action chunk reached by the teacher's full integration. 
$\frac{x_{1} - x_{\tau}}{1 - \tau}$ is the target velocity, representing the straight line direction required to solve a two-step jump. 
By minimizing this regression loss, the student model learns to internalize the straightened flow enabling the robot to execute complex tasks using only two function evaluations during inference.

\subsubsection{Classifier-Free Guidance (CFG)}
We optionally apply CFG \cite{ho2022cfg} to trade off instruction adherence and a learned interaction prior at inference time. 
To enable instruction-free predictions, we train the policy with instruction masking: during training, the instruction is replaced with a learned null instruction token with probability $p_{\mathrm{masked}}$.

We obtain two predictions: a conditional prediction $v_{\mathrm{cond}}$ with the instruction, and an instruction-free prediction $v_{\mathrm{uncond}}$ by replacing the instruction with a null token.
We then form a guided velocity field by linearly combining these two predictions:
\begin{equation}
    v_{\mathrm{cfg}} = v_{\mathrm{uncond}} + w \cdot (v_{\mathrm{cond}} - v_{\mathrm{uncond}})
\end{equation}
where $w$ controls the strength of guidance.
We treat $w$ as an operational knob controlling how strongly the instruction biases the prior.
In practice, strong guidance can make the policy more proactive in following the instruction, which can cause overly aggressive motion when $w \gg 1$. 

To stabilize, we apply a guidance rescaling step after the affine combination.
Concretely, we match the norm of the guided output to the norm of the conditional output \cite{lin2024common}:
\begin{equation}
    v_{\mathrm{rescaled}} = v_{\mathrm{cfg}} \cdot \frac{\Vert v_{\mathrm{cond}} \Vert}{\Vert v_{\mathrm{cfg}} \Vert + \varepsilon}
\end{equation}
with $\varepsilon = 1 \times 10^{-6}$ for numerical stability. 

\subsubsection{High-Frequency Control}
We reinvest the latency reduction from distillation not merely to idle the processor, but to unlock a more responsive control regime.
By reducing inference time, we can afford to shorten the execution horizon and re-infer more frequently under the same real-time budget.
This ``high-frequency feedback'' design directly counters the accumulation of drift and timing errors, boosting MTBI in dynamic real-world settings.
This design is especially effective when ESPADA is applied: because downsampling reduces the temporal density of demonstrations, we correspondingly shorten the action chunk horizon, which we find generally improves performance in practice.
Finally, we run the full perception-to-action loop on-device to eliminate network latency and ensure predictable response times. 

\subsection{Data Strategy: From Play to Task}
\subsubsection{Data collection interfaces}
To support both unstructured play and deployment-aligned cyclic demonstrations, we built a unified data collection pipeline with multiple interfaces: (1) Universal Data Device (UDD), a modular robot-free device that mounts the deployment gripper to minimize embodiment mismatch; (2) Meta Quest hand-tracking for upper-body teleoperation (retargeted to either hands or an ``easy-gripper'' abstraction); and (3) manufacturer-provided leader-arm teleoperation for robot-native demonstrations. 
See Appendix~\ref{app:data_interfaces} for detailed data collection methods.

\subsubsection{Play Dataset: Building a Robust Prior}
While task demonstrations provide the optimal trajectory for success, they fundamentally lack coverage of the state space required for recovery. 
In contrast, play data consists of unstructured teleoperated interactions and naturally captures a diverse range of behaviors, including regrasping, repositioning, and local corrections~\cite{lynch2020learninglatentplansfromplay, cui2023play_to_policy}.
Leveraging this, we collect a large-scale play corpus to serve as a foundation for general manipulation priors. 
We do not treat play data as a substitute for task data.
Instead, it expands coverage around the task manifold and increases exposure to recovery-adjacent behavior. 
By pre-training on this language-unconditioned play data, the action expert acquires a robust, task-agnostic understanding of physical interaction, captured by the unconditional prior. 

\subsubsection{Cyclic Task Dataset: Aligning for Continuity}
Standard imitation learning datasets typically consist of isolated, ``success-then-stop'' episodes initiated from curated initial states~\cite{embodimentcollaboration2025openxembodimentroboticlearning}.
We refer to this as \textbf{non-cyclic data}. 
As shown in Section~\ref{sec:experiments}, policies trained on non-cyclic data struggle under continuous repetition because they do not learn to handle the state drift across cycles.
To address this, we introduce a cyclic trajectory dataset formulation. 
Instead of resetting after each success, we collect demonstrations as continuous streams of repeated task execution. 
This exposes the model to the transition between cycles, connecting the end of one iteration to the beginning of the next, and to the drift that accumulates over time. We refer to this collection of continuous, reset-free demonstrations as the \textbf{cyclic data}.
For task-specific alignment, we curate a focused cyclic dataset totaling approximately 10 hours for the target workflow. 
This dataset serves two purposes. First, it is the sole training corpus for our baseline comparisons, establishing performance without play data priors. Second, it is the fine-tuning target for our method. By post-training on these demonstrations, we project the broad, task-agnostic motion prior acquired from play data onto the precise manifold of the target task, ensuring that the policy retains high-level goal adherence while benefiting from low-level robust behavior.

\subsubsection{Spatially Aware Downsampling (ESPADA)}
To increase physical execution speed, we apply \textbf{Spatially Aware Downsampling (ESPADA)}~\cite{kim2025espada} to demonstrations. 
Uniform downsampling can alias short, contact-rich interactions.
ESPADA avoids this by using a VLM-LLM pipeline to semantically segment trajectories into ``casual'' (transit) and ``precision'' (contact-rich) phases based on 3D gripper-object spatial relations.
We then apply selective acceleration via a replicate-before-downsample strategy, aggressively compressing casual segments while preserving high-frequency control in precision phases. 
To maintain policy stability under acceleration, we enforce geometric consistency by rescaling the action chunk horizon $H^{\prime} = \lceil \frac{H}{N} \rceil$ with the downsampling factor $N$ such that the spatial displacement matches:
\begin{equation}
\sum_{h=0}^{H^{\prime}-1} \Vert \Delta x_{t+h}^{\prime} \Vert \approx \sum_{h=0}^{H-1} \Vert \Delta x_{t+h} \Vert,
\end{equation}
where $\Delta x$ represents the original end-effector displacement, and $\Delta x^{\prime}$ represents the accelerated version. 
This ensures that the policy learns consistent spatial action distributions even when temporal density varies, mitigating slow conservative motion without sacrificing manipulation success.

\section{Experiments}
\label{sec:experiments}
To evaluate our deployment metrics, we conduct 1-hour continuous-run experiments both in simulation and in the real world. 

\subsection{Models and Baselines}
We benchmark Habilis-$\beta$ against two strong publicly available VLA architectures.
To ensure a fair comparison of architectural and methodological capabilities, all models are evaluated under the identical 1-hour continuous-run protocol.

\begin{itemize}
    \item \textbf{Habilis-$\boldsymbol{\beta}$ (Ours):} We follow all aforementioned procedures for training and inference.
    \item \textbf{$\pi_{0.5}$:} We fine-tune the publicly available $\pi_{0.5}$ checkpoint~\cite{pi05} under two training conditions: one following the standard dataset protocol and the other using our Play+Cyclic multi-task dataset.
    \item GR00T N1.5: Similarly, we fine-tune the official GR00T N1.5 checkpoint~\cite{grootn15} under two training conditions. 
\end{itemize}

All models are trained in a multi-task setting, where a single policy handles multiple tasks.

\subsection{Simulation Experiments}
\subsubsection{Simulation Setup}
We leverage the RoboTwin 2.0 \cite{chen2025robotwin} simulation suite equipped with an Aloha bimanual manipulator \cite{zhao2023aloha} as our primary testbed. To bridge the gap between episodic simulation and continuous real-world operation, we re-engineered the standard evaluation environments to support long-horizon continuous task execution.

\begin{figure}[tbp]
    \centering
    \includegraphics[width=\linewidth]{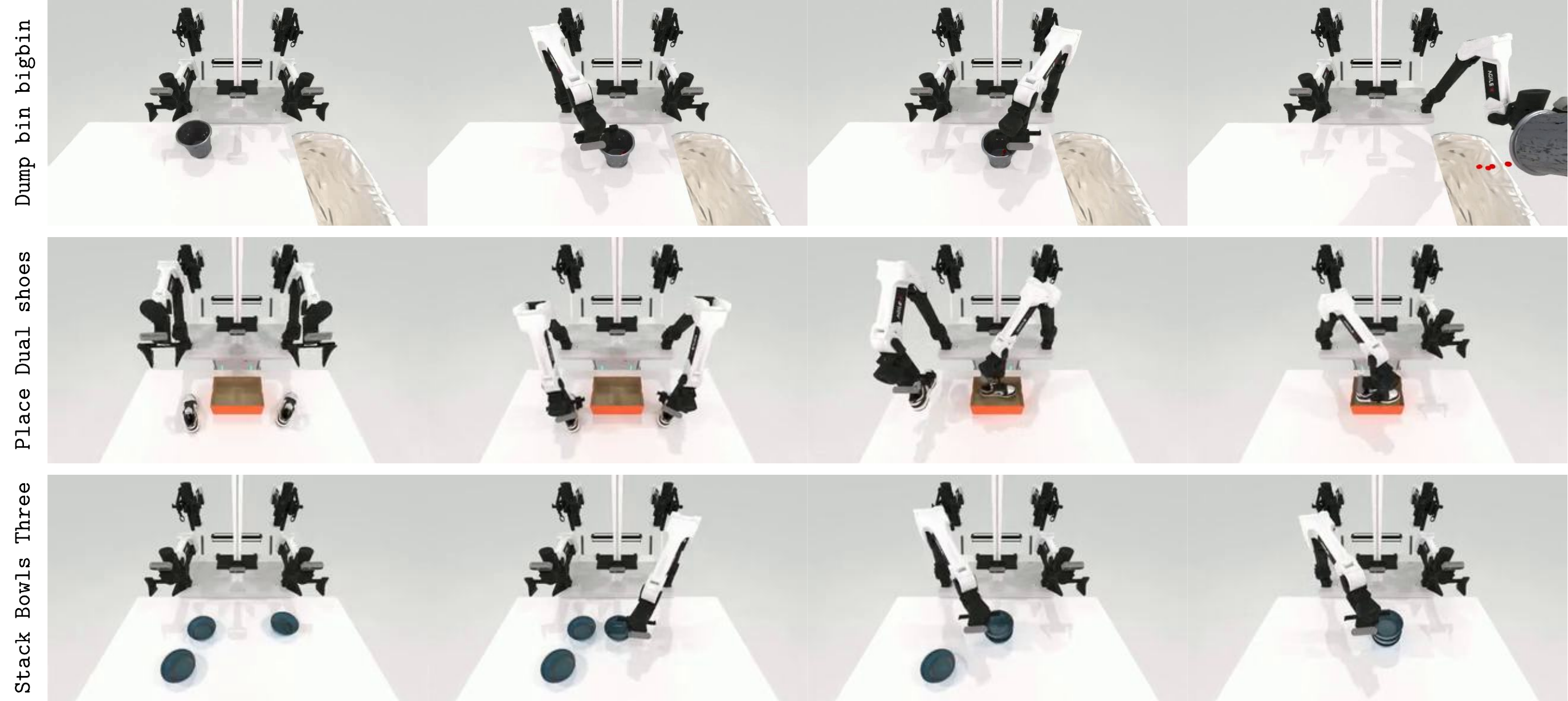}
    \caption{\textbf{Simulation Task Setup.} Simulation tasks used in our continuous-run benchmark are (top to bottom): \texttt{Dump Bin Bigbin} (DBB), \texttt{Place Dual Shoes} (PDS), and \texttt{Stack Bowls Three} (SBT). Detailed task procedures are described in the \textit{Simulation Task} paragraph below.}
    \label{fig:sim_setup}
\end{figure}

\paragraph{Simulation Tasks.}
We evaluate on three simulation tasks: \texttt{Dump Bin Bigbin} (DBB), \texttt{Place Dual Shoes} (PDS), and \texttt{Stack Bowls Three} (SBT), in the same top-to-bottom order shown in Figure~\ref{fig:sim_setup}. 

\texttt{Dump Bin Bigbin} tests long-horizon bimanual coordination with a multi-stage sequence: the agent first approaches and secures a stable grasp on one of five table bin variants, then reorients and transports the bin toward the target area; depending on the spawn location, it may include an additional repositioning step to move the object into a region that is reachable by the left hand, performs a hand-to-hand transfer when direct reach is constrained, and finally tilts the bin to dump all contents into the large destination bin.

\texttt{Place Dual Shoes} evaluates precise object placement through an ordered two-object routine: the policy identifies left/right shoe instances, performs a coordinated bimanual pickup in which both hands grasp the two shoes simultaneously when feasible, 
aligns its orientation to the designated placement frame, places the first shoe, and then repeats the same sequence for the second shoe while preserving pair-level spatial consistency.

\texttt{Stack Bowls Three} evaluates compositional manipulation via progressive stacking: the agent picks a base bowl and establishes a stable first placement, retrieves the second bowl and aligns it concentrically on top of the base, then places the third bowl to complete the stack while continuously correcting pose errors to prevent toppling.

Across all tasks, object instances and initial poses are randomized per trial to measure robustness and generalization under continuous operation.

\paragraph{Continuous-run Evaluation Protocol.}
We subject the policy to a continuous 1-hour autonomous cycle. The evaluation protocol follows a strict success-or-reset rule with task-specific time limits: for \texttt{Dump Bin Bigbin} and \texttt{Place Dual Shoes}, a trial is deemed successful if completed within 15 seconds, whereas \texttt{Stack Bowls Three} uses a 30-second limit. Upon success, the environment immediately re-spawns a new target, allowing the agent to transition seamlessly to the next trial. Failure to complete a trial within its time limit is classified as a timeout state, triggering a forced reset (intervention).

\begin{table}[b]
    \centering
    \small
    \caption{\textbf{Simulation Deployment Performance.} We report results for policies trained and evaluated on three simulated tasks: \texttt{Dump Bin Bigbin}, \texttt{Place Dual Shoes}, and \texttt{Stack Bowls Three}. For each task, we compute TPH, MTBI, and Success Rate under the continuous-run protocol, and report the average across the three tasks.} 
    \setlength{\tabcolsep}{10pt}
    \begin{tabular}{lcccc}
        \toprule
        \textbf{Method} & \textbf{Data Regime} & \textbf{TPH} & \textbf{MTBI(s)} & \textbf{Success Rate(\%)} \\
        \midrule
        $\pi_{0.5}$ & Normal & 120.5 & 30.5 & 47.8 \\
        $\pi_{0.5}$ & Play + Cyclic & 207.1 & 49.9 & 69.3 \\
        GR00T N1.5 & Normal & 1.2 & 20.4 & 0.5 \\
        GR00T N1.5 & Play + Cyclic & 5.0 & 20.6 & 2.1 \\
        \midrule
        \textbf{Habilis-$\boldsymbol{\beta}$ (w/o ESPADA)} & Play + Cyclic & 250.5 & \textbf{72.4} & 76.8 \\
        \textbf{Habilis-$\boldsymbol{\beta}$} & Play + Cyclic + ESPADA & \textbf{572.6} & 39.2 & \textbf{78.5} \\
        \bottomrule
    \end{tabular}
    \label{tab:sim_results}
\end{table}

\subsubsection{Simulation Results}
We compare Habilis-$\beta$ against the $\pi_{0.5}$, GR00T N1.5 baselines. We report metrics averaged over three tasks, each evaluated for 1-hour across three evaluation seeds.
TPH, MTBI, and success rate are computed per run and then averaged. Our empirical results are summarized in Table~\ref{tab:sim_results}. Compared to $\pi_{0.5}$, Habilis-$\beta$ increases TPH from $120.5$ to $572.6$, improves MTBI from $30.5$ to $39.2$, and raises success rate from $47.8\%$ to $78.5\%$.

\begin{figure}[H]
    \centering
    \includegraphics[width=\linewidth]{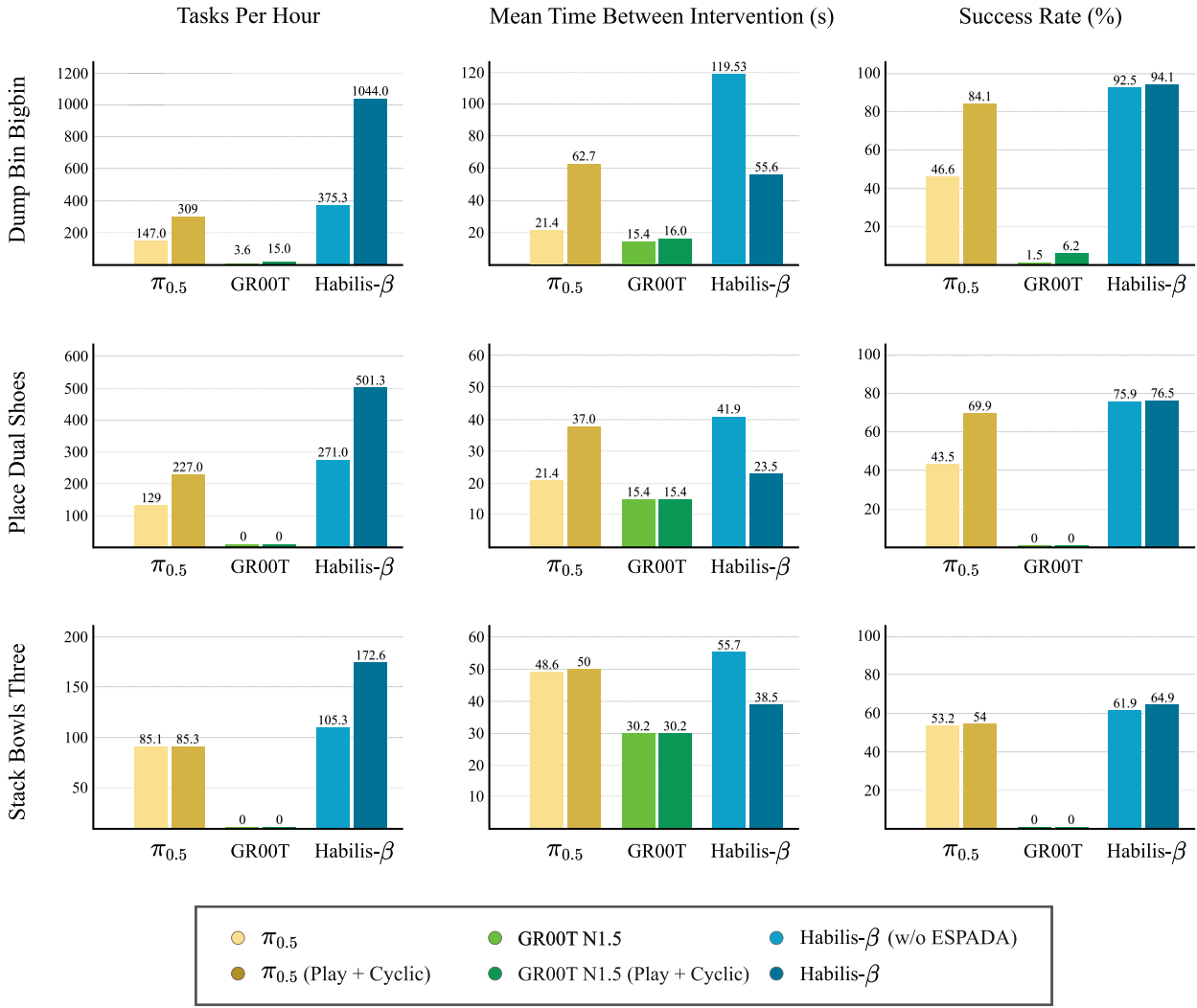}
    \caption{\textbf{Simulation Per-Task Results.} We break down the continuous-run performance metrics (TPH, MTBI, and Success Rate) for each simulation task. Habilis-$\beta$ consistently outperforms baselines across all tasks, with ESPADA significantly boosting throughput (TPH) while maintaining high success rates.}
    \label{fig:sim_per_task}
\end{figure}

\begin{figure}[tb]
    \centering
    \includegraphics[width=\linewidth]{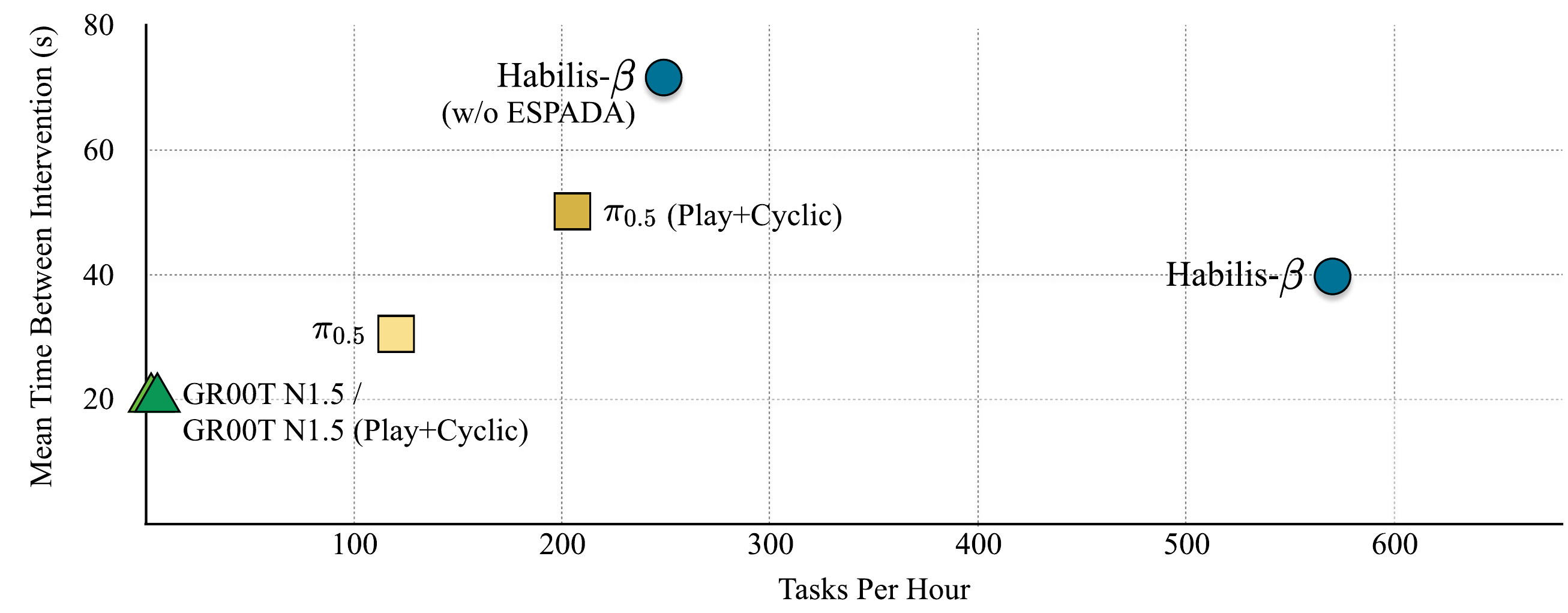}
    \caption{\textbf{Simulation Productivity-Reliability Plane.} We plot the deployment performance of different methods in simulation. Habilis-$\beta$ (w/o ESPADA) achieves the highest reliability (MTBI), while the full Habilis-$\beta$ dramatically increases throughput (TPH), offering a configurable trade-off for different operational needs.}
    \label{fig:sim_prp}
\end{figure}

A per-task comparison across methods is shown in Figure~\ref{fig:sim_per_task}.
Across all tasks, Habilis-$\beta$ achieves the highest TPH and success rate.
However, applying ESPADA leads to a consistent reduction in MTBI.
As illustrated in Figure~\ref{fig:sim_prp}, this reflects a clear productivity–reliability trade-off. 
Training with ESPADA substantially accelerates execution, yielding a large gain in productivity (TPH) while largely preserving success rate. 
However, this speedup comes with increased intervention frequency, which reduces MTBI compared to the non-ESPADA counterpart.
Intuitively, executing successful trials faster can shift the composition of outcomes: for a fixed evaluation horizon and comparable success rate, more successes per hour also imply more opportunities to encounter (time-limited) failures, increasing the share of timeout episodes that trigger interventions.
These results suggest that ESPADA primarily shifts the operating point along the PRP. 
It improves productivity by enabling faster motion, but can make the system more prone to timeouts unless paired with additional robustness mechanisms.

\subsubsection{Standard RoboTwin 2.0 Single-Trial Evaluation}
To validate the fundamental manipulation capability of Habilis-$\beta$ against state-of-the-art methods, we also conduct evaluations using the standard RoboTwin 2.0 benchmark\cite{chen2025robotwin}. 
Unlike our continuous-run setting, this protocol measures the success rate over 100 independent single-trial episodes with curated resets. 
We compare Habilis-$\beta$ against the task-wise top-performing models on the RoboTwin 2.0 leaderboard: DP3~\cite{Ze2024DP3} for \texttt{Dump Bin Bigbin}, and $\pi_0$~\cite{pi2024pizero} for \texttt{Place Dual Shoes} and \texttt{Stack Bowls Three}.

\begin{figure}[h]
    \centering
    \includegraphics[width=\linewidth]{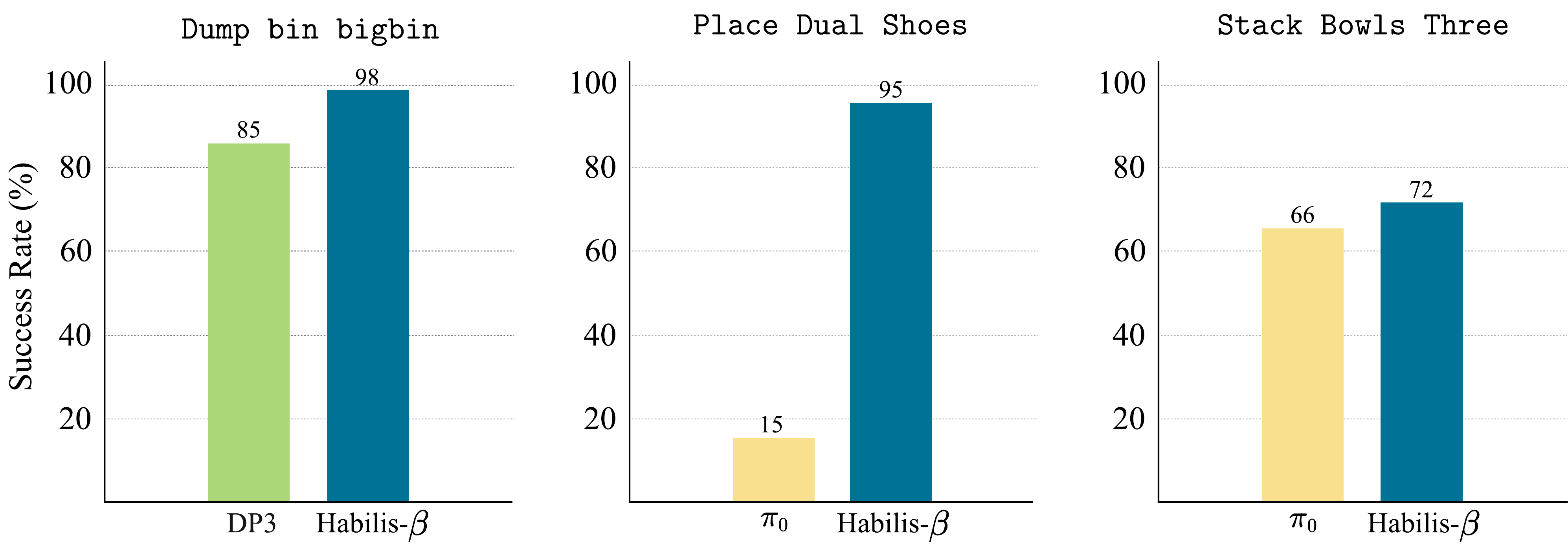}
    \caption{\textbf{Standard RoboTwin 2.0 Benchmark Results.} Success rates are measured over 100 independent episodes with curated resets under the standard RoboTwin 2.0 benchmark. Habilis-$\beta$ results are averaged across three evaluation runs to reduce seed variance.}
    \label{fig:sim_leaderboard}
\end{figure}

As shown in Figure~\ref{fig:sim_leaderboard}, Habilis-$\beta$ also achieves strong performance on the single-trial benchmark, ranking highest on the leaderboard and outperforming these baselines.

\newpage

\subsection{Real-world Experiments}
\subsubsection{Real-world Setup}
\begin{wrapfigure}{r}{0.25\textwidth}
    \centering
    \vspace{-0.4cm}
    \includegraphics[width=\linewidth]{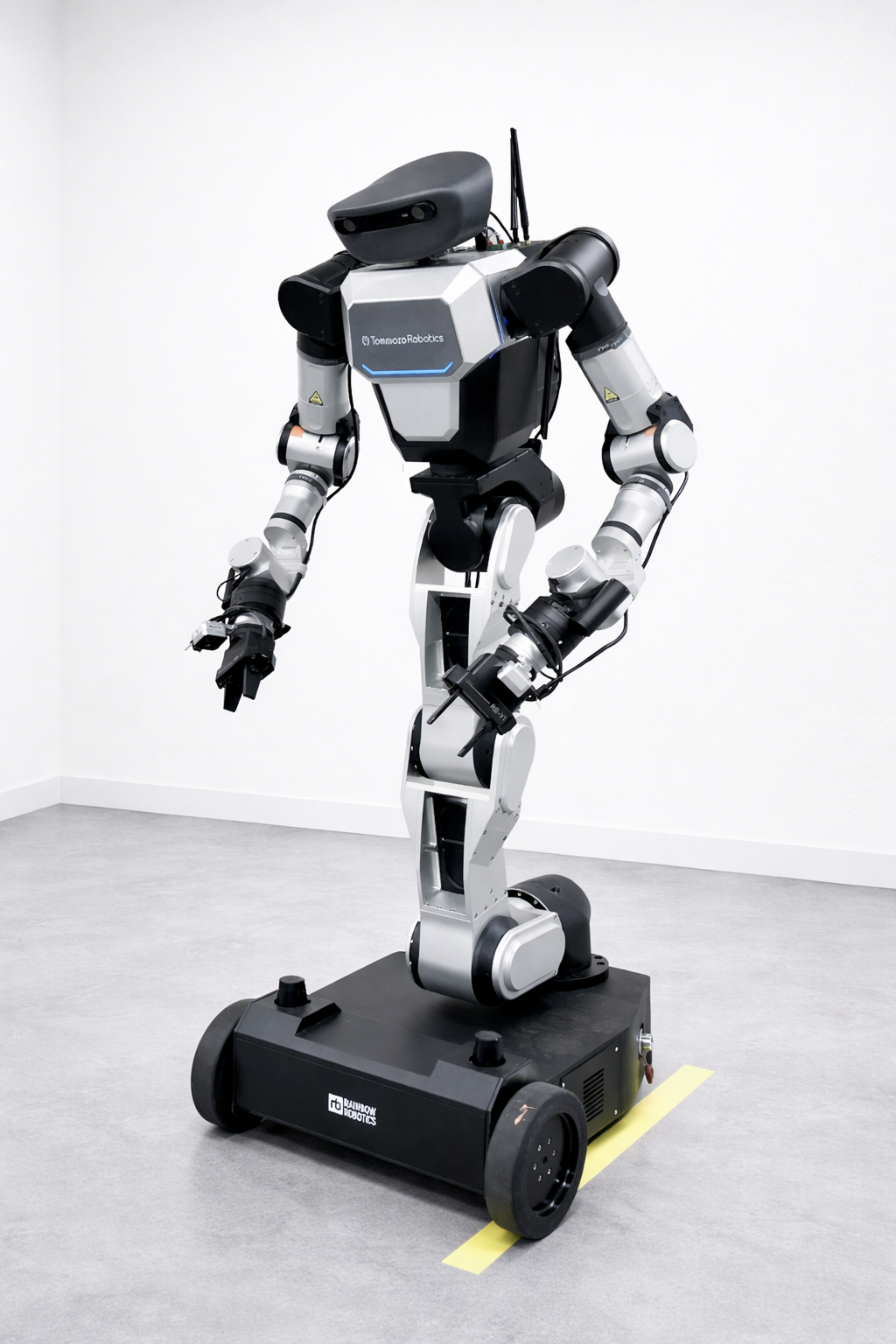}
    \caption{\textbf{RB-Y1.} We used a humanoid platform in real-world experiments.}
    \label{fig:robot_platform}
    \vspace{-1.6cm}
\end{wrapfigure}

We evaluate on an RB-Y1 humanoid robot in a repetitive industrial-style workflow. One of the target logistics tasks requires the humanoid to perform \texttt{Dual-Bin Conveyor Packing (DBCP)}: the robot retrieves an object from one of two source bins, establishes a stable bimanual grasp, and places the item into an open box moving on a conveyor, requiring both precise timing and collision-free insertion. 

\paragraph{Robot Platform.}
We utilize the RB-Y1 humanoid platform developed by Rainbow Robotics (Figure~\ref{fig:robot_platform}). This bimanual system is equipped with a multi-camera setup for robust perception: a ZED 2i stereo camera mounted on the head for global scene understanding, and Intel RealSense D405 cameras mounted on each wrist for fine-grained manipulation feedback. The end-effectors are two-finger parallel grippers designed for versatile grasping. For on-device inference, we run the full perception-to-action stack on an NVIDIA Jetson Orin.

\begin{figure}[tbp]
    \centering
    \includegraphics[width=\linewidth]{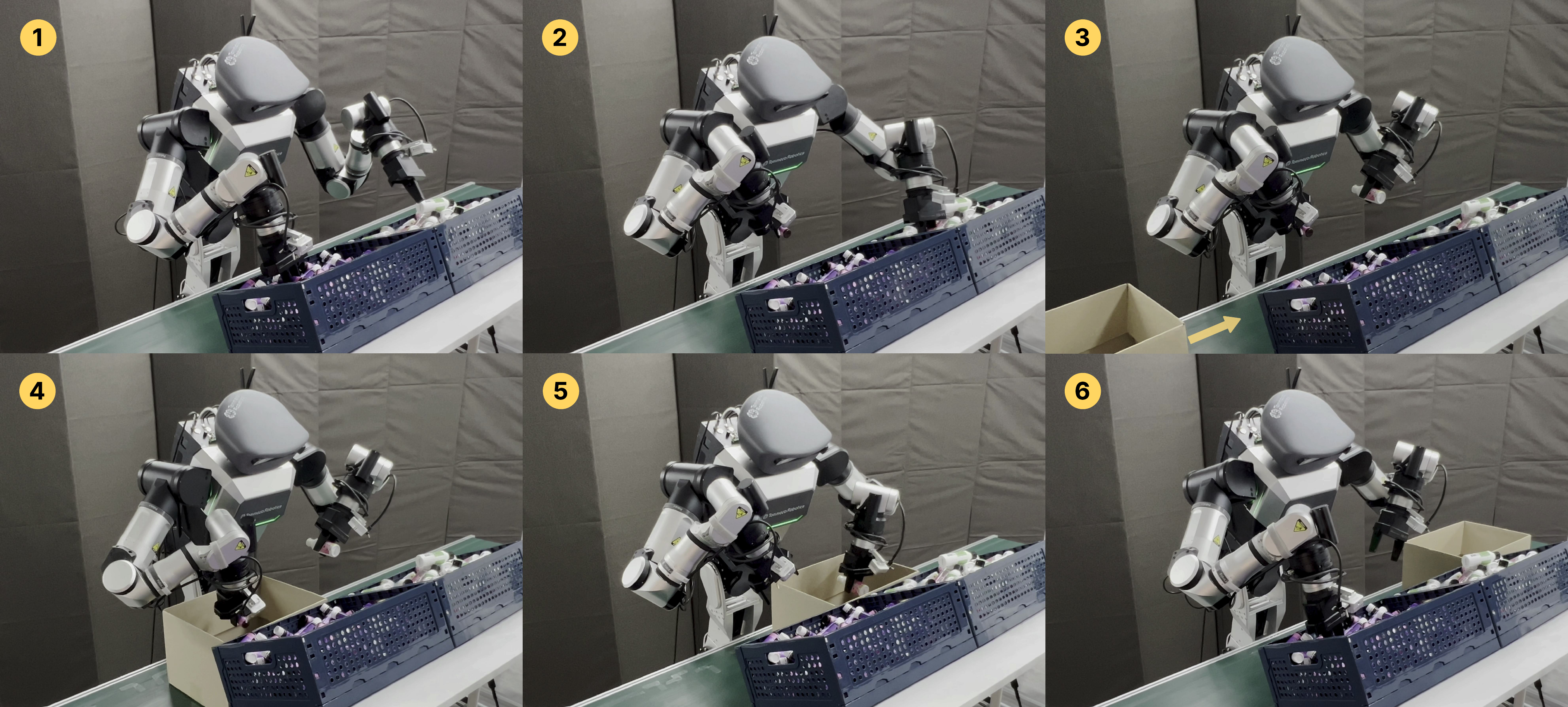}
    \caption{\textbf{Real-world Task Setup.} We conduct continuous sequential execution of the \texttt{Dual-Bin Conveyor Packing} task. The workflow involves: (1) picking an item with the right hand, (2) picking an item with the left hand, (3) waiting for the conveyor, (4) dropping the right-hand object at the correct timing, (5) dropping the left-hand object at the correct timing, and (6) immediately transitioning to the next cycle (picking with the right hand). This setup challenges the policy to maintain precise timing and coordination over long-duration runs.}
    \label{fig:real_task_setup}
\end{figure}

\paragraph{Task Specification:} \texttt{Dual-Bin Conveyor Packing}.
We evaluate our system on a high-frequency industrial workflow requiring precise bimanual coordination and dynamic temporal synchronization. As shown in Figure~\ref{fig:real_task_setup}, the task cycle is structured into three phases: (1) \textit{Right-hand Pick}, where the robot grasps an object from the right supply bin; (2) \textit{Left-hand Pick}, where the left arm retrieves an object from the left bin; and (3) \textit{Dynamic Placement}, where the agent must synchronize with a moving conveyor belt to accurately deposit both items into a passing target box. 
This task serves as a rigorous stress test for handling external timing constraints and maintaining throughput under continuous operation. The supply bins contain over 100 objects piled in random spatial distributions. Each trial alters the configuration of objects due to physical interactions, creating a dynamic and cluttered environment where precise grasping requires robust adaptation to varying object poses and occlusions. Measured TPH metric reflects not only the robot's execution speed but also its ability to consistently meet these external synchronization windows without missing a cycle.

\paragraph{Definition of Intervention.}
Unlike the simulation setting, which relies solely on timeouts, real-world interventions are categorized into two distinct failure modes: \textit{Abort} (operator-initiated termination in exceptional states requiring external recovery, e.g., an E-stop) and \textit{Timeout} (failure to complete the task within the allocated window despite continued activity). For instance, if the robot picks an incorrect number of objects, it must autonomously recognize the mistake and recover within the allotted window; otherwise, the episode cannot meet the success criteria in time and is recorded as a \textit{Timeout}, which we count as an intervention.

\paragraph{Habilis Agent.} We use \textbf{Habilis Agent} as a high-level monitoring console and logging module. While it supports broader functionality (e.g., workflow specification and supervisory control), in this work it serves to mediate interventions and provide structured, real-time logs for a fair, isolated comparison of VLA policies. When the operator determines that the current trial has entered a \textit{Timeout} or \textit{Abort} condition, they trigger a reset by clicking the corresponding action in the Habilis Agent UI; the intervention type is then automatically recorded, and the robot/environment is reset before the next trial.

\paragraph{Continuous-run Evaluation Protocol.}
Following the simulation protocol, we conduct a continuous 1-hour evaluation in the real-world environment. To ensure rigorous and fair assessment of long-term robustness, we standardize the operational maintenance procedures:
\begin{itemize}
    \item \textbf{Initial Conditions:} Each run begins with fully stocked supply bins.
    \item \textbf{Intervention-Triggered Reset:} Recognizing that grasping difficulty is highly sensitive to object distribution, we implement a standardized reset rule. If the system triggers 5 consecutive interventions, we manually shuffle the supply bins to resolve potentially adversarial clutter.
    \item \textbf{Replenishment Logic:} As bins deplete, grasping difficulty naturally increases. To maintain a consistent task challenge, if the fill level drops below 10\% and the system encounters 3 consecutive failures, we replenish the supply bins before resuming the run.
\end{itemize}

\subsubsection{Real-world Experimental Results}
We quantify the deployment readiness of each method by measuring its Productivity (TPH) and Reliability (MTBI) under the standardized 1-hour continuous protocol.

\begin{table}[h]
    \centering
    \caption{\textbf{Real-world Deployment Performance.} We compare Habilis-$\beta$ against baselines under a 1-hour continuous-run protocol. While baselines struggle with drift, Habilis-$\beta$ yields a $6.53\times$ improvement in productivity (TPH) and $2.98 \times$ reliability (MTBI) compared to the $\pi_{0.5}$ baseline.}
    \label{tab:realworld_results}
    \resizebox{0.8\columnwidth}{!}{%
    \begin{tabular}{lcccc}
        \toprule
        \textbf{Method} & \textbf{Data Regime} & \textbf{TPH} & \textbf{MTBI(s)} & \textbf{Success Rate(\%)} \\
        \midrule
        $\pi_{0.5}$ & Normal & 19 & 46.1 & 19.6 \\
        $\pi_{0.5}$ & Play + Cyclic & 79 & 94.6 & 67.5 \\
        GR00T N1.5 & Normal & 33 & 53.7 & 33.0 \\
        GR00T N1.5 & Play + Cyclic & 63 & 78.2 & 57.7 \\
        \midrule
        \textbf{Habilis-$\boldsymbol{\beta}$} & \textbf{Play + Cyclic + ESPADA} & \textbf{124} & \textbf{137.4} & \textbf{82.7} \\
        \bottomrule
    \end{tabular}%
    }
\end{table}

\begin{figure}[tbp]
    \centering
    \includegraphics[width=\linewidth]{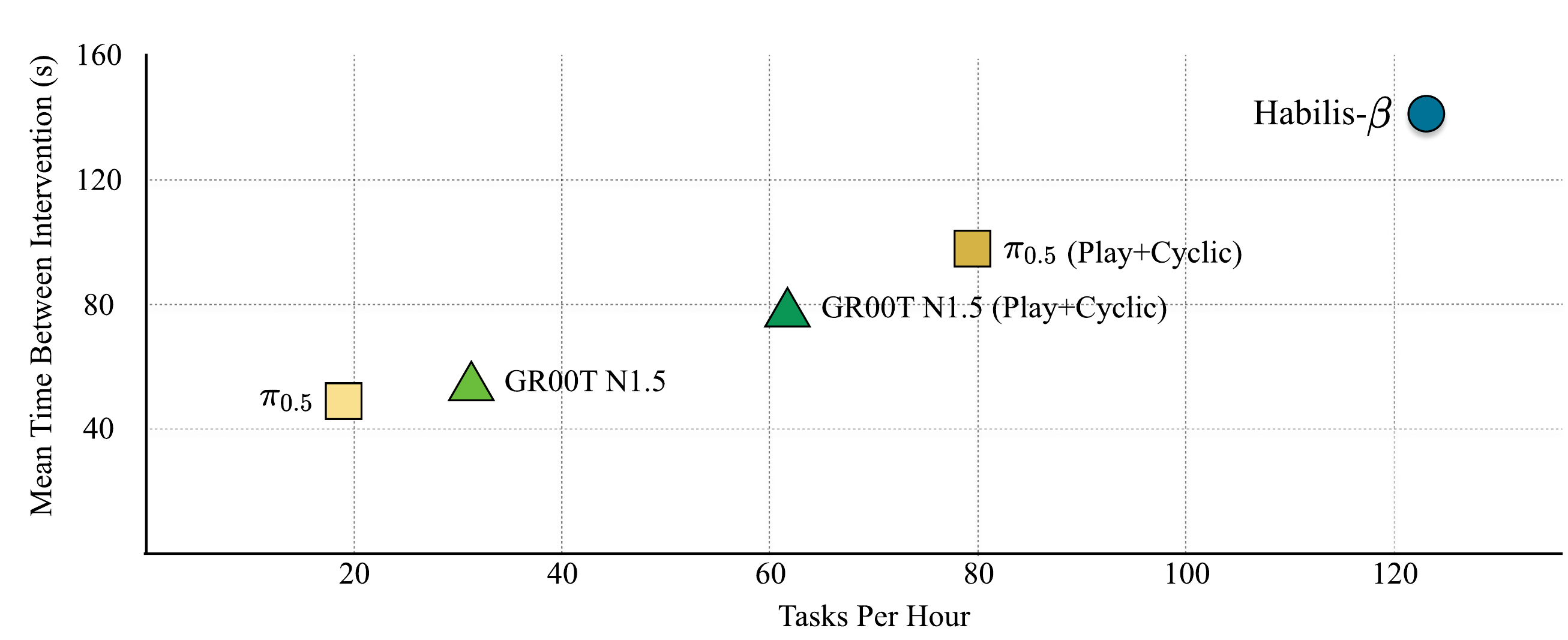}
    \caption{\textbf{Real-world Productivity-Reliability Plane.} We visualize the performance landscape, where the ideal system resides in the top-right quadrant (high throughput, high robustness).}
    \label{fig:deployment_frontier}
\end{figure}

\begin{figure}[tbp]
    \centering
    \includegraphics[width=\linewidth]{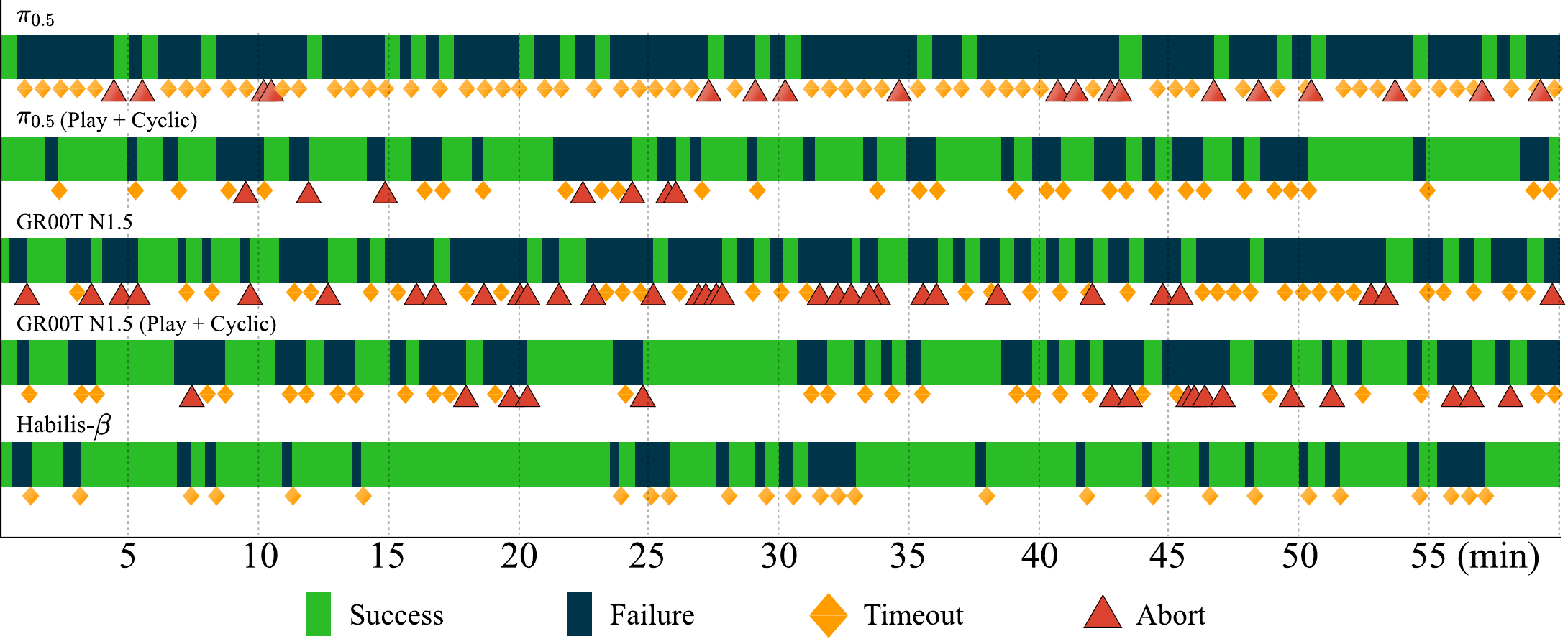}
    \caption{\textbf{Real-World Continuous Execution Timelines.} We show 1-hour continuous-run execution logs for each method. Green segments indicate successful task completions, whereas dark blue segments indicate failed attempts that do not complete within the allowed window or are terminated by the operator, and thus trigger an intervention. Markers denote intervention categories: orange diamonds for \textit{Timeout} and red triangles for \textit{Abort}. Compared to baselines, Habilis-$\beta$ achieves a higher density of successes with fewer interventions, supporting its long-lasting robustness.}
    \label{fig:real_timelines}
\end{figure}

\paragraph{The Productivity-Reliability Plane.}
Table~\ref{tab:realworld_results} and Figure~\ref{fig:deployment_frontier} summarize real-world performance, while Figure~\ref{fig:real_timelines} provides per-cycle execution timelines and intervention events over the same 1-hour runs.
With standard fine-tuning, both $\pi_{0.5}$ and GR00T N1.5 are strongly productivity-constrained ($19–33$ TPH) and achieve only moderate reliability ($46.1–53.7$s MTBI).
Moving to the Play+Cyclic data regime improves both throughput and reliability for both baselines, indicating that our data formulation provides a broadly useful training signal beyond our model.
However, Habilis-$\beta$ further advances the operating point to the top-right of the PRP, achieving $124$ TPH and $137.4$s MTBI, along with the highest success rate ($82.7\%$).

%%%%%%%%%%%%%%%%%%%%%%%%%%%%%%%%%%%%%%%%%%%%%%%%%%%%%%%%%%%%%%%%%%%%%%%%%%%%%%%
\section{Discussion}
While Habilis-$\beta$ achieves strong performance on real-world workflows, its overall generalization remains limited and has not yet reached human-level task accuracy. We expect this gap to narrow with continued scaling and engineering progress, including larger and more diverse data, improved robot hardware, and advances in foundation-model training.

Beyond scaling, our experiments highlight three practical considerations. First, the PRP view makes explicit that productivity (TPH), reliability (MTBI), and success rate are tightly coupled under a fixed operating horizon. Notably, higher productivity and success rates do not guarantee higher reliability: as the system attempts more cycles per hour, the absolute frequency of interventions may rise, effectively reducing MTBI. This aligns with industrial requirements prioritizing sustained operation, differing from conventional single-trial evaluation protocols.

Second, regarding ESPADA, while it reduces redundant motion, its semantic segmentation is mediated by an LLM. Consequently, the same demonstration may produce slightly different downsampled trajectories across runs, introducing variance that can complicate reproducibility.

Third, using CFG requires learning a strong instruction-unconditioned prior, which remains challenging given the limited scale of current robot datasets. Empirically, the optimal guidance scale varies across tasks; in some settings, disabling CFG yields the best performance. We therefore view CFG as a deployment-time parameter requiring task-specific tuning rather than a universally optimal default.

\section{Future Work}
Future work will focus on three directions that we believe will be important in the long run. First, our current robot-free interfaces provide limited supervision for dexterous embodiments; we plan to extend these to improve action/observation alignment for five-finger manipulation. Second, the system lacks tactile feedback; we will incorporate tactile inputs to enhance robustness in contact-rich phases and under partial observability. Finally, to address the static nature of offline training, we will explore safe online adaptation and continual learning, allowing the system to correct recurring failure modes discovered during long-term deployment.

%%%%%%%%%%%%%%%%%%%%%%%%%%%%%%%%%%%%%%%%%%%%%%%%%%%%%%%%%%%%%%%%%%%%%%%%%%%%%%%%%%%%%%%

\begingroup
\small
\bibliographystyle{plain}
\bibliography{references}
\endgroup

\clearpage
\appendix
% TODO: include detailed protocols, additional plots, and infrastructure details if needed.

\section{Contributions and Acknowledgements}
\label{sec:contrib}
\textbf{Core Contributors.} Jesoon Kang, Taegeon Park, Jisu An, Soo Min Kimm, Jaejoon Kim \\
\textbf{Contributors.} Jinu Pahk, Byungju Kim, Junseok Lee, Namheon Baek, Sungwan Ha, Hojun Baek, Eduardo Ayerve Cruz, Wontae Kim, Junghyeon Choi, Yousuk Lee, Joonmo Han, Sunghyun Cho, Sunghyun Kwon, Soyoung Lee \\
\textbf{Advisors.} Jun Ki Lee, Seung-Joon Yi, Byoung-Tak Zhang \\
\textbf{Project Lead.} Theo Taeyeong Kim

We thank the data collection team for their efforts in collecting and curating the dataset. Special thanks go to Soo Beom Hwang, Jaehyun Jeong, and Jaehun Heo for their contributions to assembling the \texttt{Dual-Bin Conveyor Packing} task dataset. We are also grateful to Robot Field Engineer Kangin Lee for his essential technical support and hardware maintenance throughout the experiments.

\section{Data Collection Interfaces}
\label{app:data_interfaces}

\paragraph{Universal Data Device.}
Universal Manipulation Interface (UMI)~\cite{chi2024umi} proposes a portable demonstration interface based on a hand-held, sensorized gripper and an aligned observation/action interface for policy learning. It emphasizes (1) portability for collecting demonstrations outside the lab, and (2) transferability across robot platforms by using inference-time latency matching and a relative-trajectory action representation.
In our setting, however, the target embodiments vary substantially across deployments (e.g., different grippers/end-effectors and mounting geometries). This made a one-size-fits-all handheld interface insufficient: even small differences in grasp geometry, actuation limits, and camera extrinsics can induce systematic distribution shifts that accumulate in long-horizon, repeated execution \cite{honerkamp2025wholebodyteleoperation}.

To address the above, we designed Universal Data Device (UDD) as a modular data collection device whose \textit{front-end is the actual robot gripper} (or a mechanically equivalent gripper module). The core design goal is to minimize the action/observation embodiment gap by anchoring demonstrations to the end-effector that will be used at deployment time, while still enabling robot-free data collection.
Concretely, UDD provides:
\begin{itemize}
    \item \textbf{Universal gripper mounting}: a quick-swap mechanical/electrical interface that allows attaching different manufacturer grippers without redesigning the rest of the device.
    \item \textbf{Pose tracking for 6D actions}: visual-inertial tracking (and/or equivalent onboard sensing) to recover end-effector pose trajectories with sufficient fidelity for fast motions and contact-rich segments.
    \item \textbf{Gripper actuation logging}: direct sensing of gripper instructions/states (e.g., width/force/trigger signals depending on the gripper), recorded in sync with pose and vision streams.
\end{itemize}
With UDD, collecting VLA trajectories no longer depends on bringing the full robot to every environment. Instead, the device preserves the deployed end-effector’s geometry and control constraints, making the collected data compatible with diverse robot embodiments via a consistent interface layer.

\paragraph{Meta Quest teleoperation via hand tracking.}
In addition to device-based demonstrations, we developed an upper-body teleoperation pipeline using Meta Quest hand tracking. The system provides time-synchronized 3D hand poses (and joint configurations) and retargets them to either (1) a hand embodiment or (2) a lightweight ``easy gripper'' abstraction in which pinch/open gestures are mapped to gripper commands. Notably, this setup captures dexterous human hand motion using only an off-the-shelf VR device, eliminating the need for expensive external motion-capture hardware, markers, or dedicated tracking infrastructure.

\paragraph{Leader-arm robot demonstrations.}
Finally, our dataset includes demonstrations collected using manufacturer provided leader-arm (master) interfaces. This provides high-precision robot-native demonstrations that complement the broader coverage and diversity of UDD/hand-tracking data.

\section{Habilis Agent}

Habilis Agent is a software system designed for defining robot workflows, integrating components (e.g., VLAs, navigation), and monitoring inference. 
In this work, we primarily utilize it as a structured logging interface for long-horizon evaluation. 
Operators use the UI to label trial states (e.g., \textit{Timeout} or \textit{Abort}) and trigger resets; the system automatically executes the reset procedure and records the intervention. 
Additionally, Habilis Agent logs a comprehensive execution timeline—tracking successes, timeouts, and aborts, to facilitate debugging and enable reliable TPH/MTBI measurement across long-duration experiments.

\section{Habilis Console}

Habilis Console is a platform for collecting robot demonstration data by connecting diverse robots and teleoperation devices, managing datasets, and training Robot Foundation Models (RFMs). Most of our data collection pipeline and model training, covering both the baselines and Habilis-$\beta$, were conducted within Habilis Console.

Further details are available at \url{https://tommoro-ai.github.io/habilis-console/}, and the platform is scheduled for an official release in March 2026.

\section{Related Works}

\subsection{Vision-Language-Action Models}
\paragraph{Foundations of Generalist Robot Policies.}
The success of foundation models in NLP and vision introduced a new paradigm for robotics: a single, high-capacity model, pre-trained on diverse data, that can solve new tasks with little or no task-specific adaptation~\cite{kawaharazuka2025vla}.
This idea has taken shape as VLA models.
Early systems like RT-1 \cite{Brohan-RSS-23} tokenized continuous actions into a discrete vocabulary, utilizing transformers to map image-text inputs to action tokens.
RT-2~\cite{pmlr-v229-zitkovich23a} then demonstrated that large pre-trained vision-language models (VLMs) can be adapted to output robot actions, leveraging broad semantic representations and language-grounded reasoning.
In parallel, scaling efforts and open-source work have produced generalist policies such as Octo~\cite{octo2024}.
More recently, approaches such as OpenVLA~\cite{kim2024openvla, openvla-oft} fine-tune VLM backbones to output robot actions directly, further strengthening the VLM-centric direction.

\paragraph{Action Generation with Diffusion and Flow Matching.}

Alongside model scaling, action generation in robot policies has increasingly relied on generative modeling.
Diffusion-based policies~\cite{chi2023diffusionpolicy, Ze2024DP3}, grounded in diffusion/score-based models~\cite{ho2020denoisingdiffusionprobabilisticmodels, song2021scorebasedgenerativemodelingstochastic}, gained early traction for their ability to represent multimodal action distributions, albeit at the cost of iterative denoising and higher inference latency.

To mitigate this computational burden, flow matching objectives were introduced as an alternative that learns a direct (often deterministic) transport from noise to actions, enabling substantially fewer inference steps~\cite{lipman2023flow_matching}.
Representative early applications in robotics include ManiFlow~\cite{yan2025maniflow} and $\pi_0$~\cite{pi2024pizero}, which demonstrated that flow-based action generation can preserve action quality while improving sampling efficiency.

More recently, these two paradigms have been adopted in foundation-scale VLA systems.
For example, GR00T N1.5 \cite{bjorck2025gr00tn1} follows a diffusion-style action generator, while $\pi_{0.5}$ \cite{pi05} builds upon flow-matching-style generation.
This progression highlights the practical trade-off between action expressivity and inference cost in deployed control loops.

However, capability gains do not automatically translate to deployment readiness.
In repetitive industrial settings, generalist models can struggle to sustain high throughput and stable operation over long runs, where both productivity and reliability are first-class constraints.

\subsection{Fast-Motion}
While VLAs have improved in semantic understanding, physical execution speed remains a major bottleneck for deployment.
In practice, many policies produce overly fine-grained, temporally dense action sequences, taking many small steps even in phases where rapid motion would be safe.
This conservative temporal behavior often reflects the granularity of teleoperation data and directly reduces throughput in repetitive workflows.

\paragraph{Data-level Speed Shaping.}
Recent efforts attempt to increase action-level speed by modifying the temporal structure of demonstrations.
Heuristics such as DemoSpeedup~\cite{guo2025demospeedup} and SAIL~\cite{arachchige2025sail} compress demonstrations or vary control rates to accelerate free-space motion.
While these methods can improve throughput, they are largely heuristic and may risk instability when contact-rich phases are compressed.

\paragraph{The Need for Semantic-Temporal Adaptivity.}
Fast-motion requires more than uniformly running at a higher speed; it requires reasoning about when to move fast.
A deployment-oriented policy should treat free-space motion differently from contact and precision phases. Free-space segments can be coarse and fast, while interaction segments must remain fine-grained and reactive.
This motivation underpins our adoption of ESPADA~\cite{kim2025espada} in Habilis-$\beta$, which adjusts temporal granularity using task-relevant semantics to accelerate transit while preserving contact-critical control.

\subsection{Long-Lasting}
\paragraph{Industrial Metrics vs. Academic Success Rates.}
In traditional manufacturing, production performance is often tracked using Overall Equipment Effectiveness (OEE) \cite{oee}, alongside reliability metrics such as Mean Time Between Failures (MTBF) \cite{mtbf}.
In contrast, much of academic VLA research emphasizes episode-level success rate under short horizons and curated resets. 
Recent long-horizon works (e.g., Long-VLA~\cite{fan2025long}, LoHoVLA~\cite{yang2025lohovla}) extend the context length, but they are not explicitly designed for long-lasting, repetitive deployment where policies must sustain performance under continual execution with minimal human intervention.

\paragraph{The ``Intervention'' Bottleneck.}
In production, the dominant cost is often not a single failure on a difficult task. 
It is the frequency of human intervention caused by drift, accumulated error, or frozen states over hours of operation. 
Yet standardized evaluation protocols that measure this ``intervention-free'' capability remain limited \cite{zhang2024vlabench, guruprasad2024benchmarkingvla, mees2022calvin}. 
To bridge this gap, we adopt Mean Time Between Intervention (MTBI). 
MTBI measures the average operating time between events that require external intervention. 
We argue that industrially viable VLA systems must incorporate priors that prevent drift and support recovery, which is a central design goal in Habilis-$\beta$.

\paragraph{Play Data.}
Large-scale play data is attractive for learning robust, low-level interaction skills because it can be collected without task segmentation, language annotation, or curated resets. 
Beyond lowering data acquisition cost, play data often includes recovery-adjacent states and diverse contact configurations that are underrepresented in success-only demonstrations. 
This matters for sustained operation under repetition. 
Prior works  \cite{lynch2020learninglatentplansfromplay, LynchSermanet2021LCIL} show how play data can be structured into reusable latents and leveraged for downstream goal-directed control, improving robustness and retry behavior even without explicit task labels.

\subsection{On-Device}
On-device execution matters because it makes end-to-end latency and reliability less sensitive to network variability and remote service dependencies.
However, running the full perception-to-action loop locally imposes tight compute and power budgets, motivating VLA models and runtimes designed explicitly for efficient inference.
Recent works explore efficiency from multiple angles, including lightweight language-conditioned policies~\cite{clip-rt} and dynamic inference mechanisms that adapt computation to the current control context~\cite{yue2024a}.
Architectural and decoding strategies such as RoboMamba~\cite{liu2024robomamba} (SSM-based backbones) and Spec-VLA~\cite{wang2025specvla} (speculative decoding) further target reduced inference latency for real-time control.

\paragraph{Rectified Flow.} 
For generative action models, deployment constraints often translate to a need to reduce iterative sampling while preserving action quality. 
Distillation and step-reduction techniques address this by compressing multi-step generation into a small number of solver steps, enabling higher-frequency re-inference within a fixed latency budget. 
Rectified flow \cite{liu2022rectified_flow} offers a framework that supports such step-reduction via trajectory rectification and related distillation-style constructions.

\subsection{Classifier-Free Guidance (CFG)}
\paragraph{Classifier-Free Guidance.} 
CFG \cite{ho2022cfg} offers an inference-time control knob by mixing conditional and unconditional predictions to adjust how strongly the model follows task conditioning versus learned priors. 
While CFG has appeared in several policy-learning settings \cite{lu2025cfgdp, chen2025modulardp, li2025lano3dp}, our use follows the traditional conditional generation setup. 

\end{document}